\documentclass{article}
\pdfoutput=1

 \usepackage[final]{neurips_2019}

\usepackage[utf8]{inputenc} 
\usepackage[T1]{fontenc}    
\usepackage{hyperref}       
\usepackage{url}            
\usepackage{booktabs}       
\usepackage{amsfonts}       
\usepackage{nicefrac}       
\usepackage{microtype}      
\usepackage{natbib}
\usepackage{multirow}
\usepackage{subcaption}
\usepackage{graphicx}
\usepackage{amsmath}
\usepackage{longtable}
\newcommand{\Lagr}{\mathcal{L}}
\newcommand{\rulesep}{\unskip\ \vrule\ }
\usepackage[title]{appendix}
\usepackage{commath}

\title{Learning Maximally Predictive Prototypes in Multiple Instance Learning}

\author{%
    Mert Yuksekgonul \\
    Department of Industrial Engineering \&\\
    Department of Computer Engineering \\
    Bogazici University, Istanbul, Turkey\\
  \texttt{mert.yuksekgonul@boun.edu.tr} \\
   \And
   Ozgur Emre Sivrikaya\\
   Department of Industrial Engineering\\
    Bogazici University, Istanbul, Turkey\\
   \texttt{ozgur.sivrikaya@boun.edu.tr} \\
   \AND
   Mustafa Gokce Baydogan\\
   Department of Industrial Engineering\\
    Bogazici University, Istanbul, Turkey\\
   \texttt{mustafa.baydogan@boun.edu.tr} \\
}

\begin{document}

\maketitle

\begin{abstract}
In this work, we propose a simple model that provides permutation invariant maximally predictive prototype generator from a given dataset, which leads to interpretability of the solution and concrete insights to the nature and the solution of a problem. Our aim is to find out prototypes in the feature space to map the collection of instances (i.e. bags) to a distance feature space and simultaneously learn a linear classifier for multiple instance learning (MIL). Our experiments on classical MIL benchmark datasets demonstrate that proposed framework is an accurate and efficient classifier compared to the existing approaches. 
\end{abstract}

\section{Introduction}

\paragraph{} Classification problems can be divided into two with respect to the labeling characteristics of the data, single instance (SI) and multiple instance (MI) problems. In single instance learning (SIL) problems, each instance is individually labeled. However, multiple instance learning (MIL) concentrates on bags of instances, not individually labeled instance data. Two different labeling characteristics of these problems can be seen in Table \ref{table:single-multi}.

\begin{table}[!htb]
    \begin{subtable}{.5\linewidth}
      \centering
    \begin{tabular}{ |c|c|c| }  
\hline
Bag  & Instance & Label \\
\hline
\multirow{2}{1em}{$B_1$}  & $X_{1}$ & \multirow{2}{1em}{1} \\ 
& $X_{2}$ & \\ 
\hline
\multirow{2}{1em}{$B_2$}  & $X_{3}$ & \multirow{2}{1em}{0}  \\ 
& $X_{4}$ &  \\
& $X_{5}$  & \\
\hline
       \end{tabular}
        \subcaption[]{MIL Problem}
    \end{subtable}%
    \begin{subtable}{.5\linewidth}
      \centering
          \begin{tabular}{ |c|c| }  
\hline
Instance & Label \\
\hline
$X_{1}$ & 1 \\ 
\hline
$X_{2}$ & 0 \\ 
\hline
$X_{3}$ & 1 \\ 
\hline
$X_{4}$ & 1 \\ 
\hline
$X_{5}$ & 0 \\ 
\hline
        \end{tabular}
\caption[]{SIL Problem}
    \end{subtable} 
 \caption{Labeling in MIL and SI Problems}
\label{table:single-multi}
\end{table}

\paragraph{} Unlike many common problems in machine learning, multiple instance learning problems do not have a fixed input size. The number of instances in a bag is often variable, hence the most widely used architectures like deep neural networks are not straightforward to apply to these problems. Mainly there are two different approaches to address this issue, first one is instance level approaches and the second one is embedding approaches. The instance level approaches try to reward a probability per each instance that exists in a bag, then apply some pooling function to probabilities to obtain the final bag probability. In the second type of approaches, an arbitrary function, often a neural network, is used to come up with an embedding for each instance, then again some pooling function is applied to aggregate information from each embedding which is fed to a classifier. In this work, we propose an embedding approach with some modifications. The idea is to find some representative prototypes in the feature space so that bags are linearly separable when they are represented as their distances to the prototypes.

\paragraph{}The rest of the paper is organized as follows: Section \ref{RelatedWork} gives an overview of the previous studies of the field. Section \ref{Method} explains learning algorithm and the solution method of the problem. Finally, results of the solution approach on various data sets and our conclusions can be found in Sections \ref{Experiment} and \ref{conclusion}.

\section{Related Work}\label{RelatedWork}
Previous research on MIL problems start with a standard assumption of the problem. Standard MIL assumption tells that if a bag has at least one positive instance, then bag's label is positive \citep{dietterich}. Later, solution approaches are proposed for problem's different variations and extensions \citep{weidmann_et_al, classifier-combining, lazy_learning}. Mainly, two different approaches are adopted for all extensions of MIL problems. First one is instance level \citep{xu_and_frank, xu, raykar_et_al} or bag level \citep{amores, zhou_and_jiang} approaches as indicated previously. However, instance level approaches have dimensionality problem and it is not always possible to solve MIL problems with instance based approaches. Bag level approaches overcome dimensionality but have a disadvantage of losing information that can be gathered from instances. Embedding approaches are developed to overcome these disadvantages \citep{cheplygina2013combining, dissimilarity, gehler}. 

\paragraph{}
Just like every other differentiable task, application of neural network based approaches to MIL problems has been drawing attention from several different domains in the recent years \citep{deepmil, deepmil2}. Well known problems have also been reformulated for this particular purpose, such as common computer vision tasks like image classification \citep{classification}, weakly supervised object detection \citep{object1, object2}; sequence predictions \citep{resource},  sentiment analysis \citep{sentiment} and sound event detection \citep{sound_event}.

\section{Method}\label{Method}

\subsection{Joint Learning of Prototypes and Classification}
\begin{figure}[!h] \label{fig:1}
  \centering
  \begin{subfigure}[b]{0.47\linewidth}
\centering
    \includegraphics[width=\linewidth]{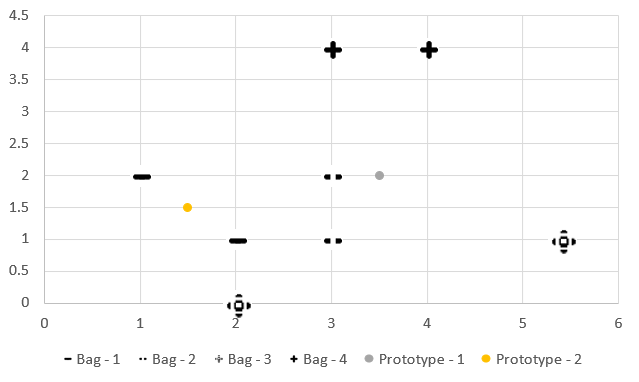}
    \caption{Bag  1  and  Bag  2  are  negative  labeled bags.  Bag 3 and Bag 4 have positive labels.  All bags and prototypes have two features.}

\label{fig:bag_a}
  \end{subfigure}
  \rulesep
  \begin{subfigure}[b]{0.47\linewidth}
    \includegraphics[width=\linewidth]{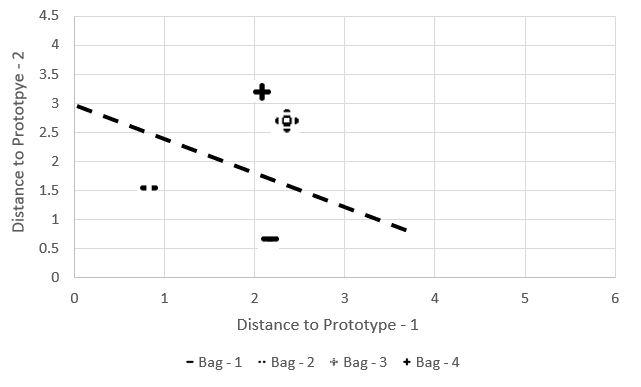}
    \caption{Distances of bags to prototypes.  A good prototype candidate separates negative and positive labeled bags. }
\label{fig:bag_b}
  \end{subfigure}
  \caption{Representation of Bags and Prototypes.}
  \label{fig:bag}
\end{figure}

\begin{figure}[!h]
  \centering
  \includegraphics[width=\linewidth]{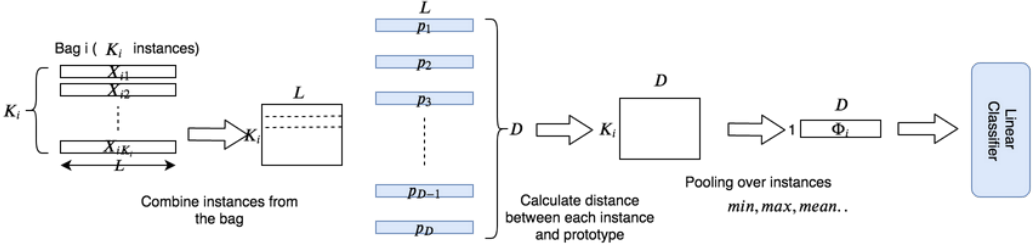}
  \caption{Overview of the architecture. Blue color shows the variables, which the model will be optimized over. $L$: Number of features in an instance, constant; $D$: Number of prototypes, hyperparameter; $K_i$: Number of instances bag i, varies between different bags}
  \label{fig:arch}
\end{figure}

The model has two main objectives: First one is learning the feature vectors or prototypes that is maximally predictive of the bag class after finding an embedding in the distance space. A simple illustration of the idea is presented in Figure \ref{fig:bag}. Suppose there are two positive and two negative bags each of which have two instances in two-dimensional feature space shown in Figure \ref{fig:bag_a}. Our aim is to identify prototypes such that the bags are linearly separable when each bag is represented by its minimum distance to each prototype. Figure \ref{fig:bag_b} represents the bags in the new feature space. In other words, proposed model is optimized over both the linear classifier parameters and the prototypes. An overview of architecture can be seen in Figure \ref{fig:arch}. Depending on the application, our proposal is flexible in generating average and maximum type of features which are famous in multiple instance learning domain \citep{bag_dissimilarity}.
\\
\paragraph{}
In terms of interpretability, \cite{deepmil} uses attention to give weights to instances in a bag and use these weights to do classification, hence in a sense one can interpret which instance contributes to the decision. One of the main contributions of our work is, we are able to extract meaningful prototypes in a dataset that represent classes. In other words, we are not only able to put meaningful weights on instances, but also we can extract the most predictive fragments in the dataset with respect to the task. After the training phase is complete, we can examine the prototypes to see what fragments are the most representative of the given class. Moreover, since we are not using a feedforward network to extract the features, but we are using the distances to each prototype as the features, we do not need a computationally expensive architecture. In other words, after calculating the distances to prototypes, we only use a simple linear layer to make a prediction.

\paragraph{}In our setting, for a given training task we choose a fixed number of prototypes, $D$, of a fixed size, $L$, to be learned, we initialize these prototypes randomly. We combine each instance of length $L$ in a given bag, which yields $K_i$(number of instances in bag i) vectors representing bag i. $K_i$ is not constant between different bags, since each bag potentially has different number of instances.  
At each training step, we calculate the distance from each instance to the prototypes to extract distance features. Given these features, the model learns a classifier to predict the bag class. 

\subsubsection{Distance Feature Extraction}
Just as in instance-level approaches in MIL problems, our model also needs to pool information that is extracted from instances in a given bag with potentially variable number of instances. To be more specific, for a given bag after the distance from each instance to each prototype is calculated, the model needs to aggregate the information before being fed into a linear classifier. These pooling operations should be differentiable to be optimized with a gradient based approach. Most basic and widely used pooling operators having these characteristics are min, mean and max operators \citep{bag_dissimilarity}. These are also intuitively informative in our case, since we have distance metrics as features, such that these should provide information about defining characteristics of an instance, assuming the existence of prototypes which are described above. After the distance extraction, we apply L2 regularization to all extracted distances, to minimize the distances. This is to ensure that the prototypes are as close to instances as possible and semantically meaningful. Our objective function for an example problem can be found in Appendix A Equation \ref{eq:objective}.
%
\subsubsection{Feature Normalization}
The distance features are prone to scale issues. This can cause problems with both gradient updates and the learning of linear classifier parameters. To overcome this, we adapt a similar approach to \cite{layernorm}, \cite{batchnorm}. In other words, for each bag, we normalize the aggregated distance vector.
Note that the information related to "shape" of the distance features will be preserved under the normalization operation that is only recentering and rescaling, which is what we aim to achieve. 
\\
A detailed formulation of the described method can be found in Appendix \ref{appendix}.
\section{Experiments}\label{Experiment}
Solution approach is tested on five MIL datasets from two categories, molecular activity prediction and image annotation. We repeat a stratified 10-fold cross-validation five times and report the average of the classification accuracy with standard error in Table \ref{tab:Results}. For all experiments, prototypes are generated randomly and logistic regression is used as default classifier. Important parameters are the number of prototypes to learn and learning rate for weights and prototypes. The model was implemented in PyTorch\citep{pytorch} and we use the same parameters for each dataset. Namely, the number of epochs is set to 100 with a minibatch size of 1. Adam optimizer from \citep{adamopt} was used. As mentioned above, we use different learning rates for the classifier and the prototypes. Regularization parameters and learning rates for each dataset can be found below.

\begin{table}[h!]
    \centering
    \resizebox{\textwidth}{!}{
    \begin{tabular}{cccccc} 
\hline
\textbf{}                                                                                    & \textbf{Musk1} & \textbf{Musk2} & \textbf{Fox}  & \textbf{Tiger} & \textbf{Elephant}  \\\hline
Learning Rate of Classifier       & 3e-5   & 4e-5   & 3e-5   & 1e-4   & 3e-5   \\
Learning Rate of Prototypes       & 9e-5 & 8e-5 & 5e-5 & 3e-5 & 9e-5 \\
$\lambda_p$ , Prototypes Regularization Parameter          & 4e-3 & 4e-3 & 4e-3 & 4e-3 & 4e-3 \\
$\lambda_d$ , Distance Regularization Parameter       & 1e-2 & 1e-2 & 1e-2 & 1e-2 & 1e-2 \\
$\lambda_d$ , Classifier Weight Regularization Parameter         & 3e-4 & 3e-4 & 3e-4 & 3e-4 & 3e-4 \\
Number of prototypes & 24 & 24 & 24 & 24 & 24 \\
\hline \\
       \end{tabular}}
    \caption{Hyperparameters for different datasets.}
    \label{tab:Results}
\end{table}

\subsection{Classification Accuracy}

Solution approach in this study outperforms or at least does as good as all other well-known methods in terms of classification accuracy. Besides this approach has much less parameters compared to a neural network, namely only prototypes and few parameters for linear classifier.

\begin{table}[h!]
    \centering
    \resizebox{\textwidth}{!}{
    \begin{tabular}{cccccc} 
\hline
\textbf{}                                                                                    & \textbf{Musk1} & \textbf{Musk2} & \textbf{Fox}  & \textbf{Tiger} & \textbf{Elephant}  \\\hline
mi-SVM \citep{andrews2003support}        & 0.874$\pm$N/A   & 0.836$\pm$N/A   & 0.582$\pm$N/A   & 0.784$\pm$N/A   & 0.822$\pm$N/A   \\
MI-Kernel\citep{positive-negative}       & 0.880$\pm$0.031 & 0.893$\pm$0.015 & 0.603$\pm$0.028 & 0.842$\pm$0.010 & 0.843$\pm$0.016 \\
EM-DD\citep{zhang_and_goldman}           & 0.849$\pm$0.044 & 0.869$\pm$0.048 & 0.609$\pm$0.045 & 0.730$\pm$0.043 & 0.771$\pm$0.043 \\
mi-Graph\citep{zhou}        & 0.889$\pm$0.033 & 0.903$\pm$0.039 & 0.620$\pm$0.044 & 0.860$\pm$0.037 & 0.869$\pm$0.035 \\
miVLAD\citep{wei2017scalable}          & 0.871$\pm$0.043 & 0.872$\pm$0.042 & 0.620$\pm$0.044 & 0.811$\pm$0.039 & 0.850$\pm$0.036 \\
miFV\citep{wei2017scalable}            & 0.909$\pm$0.040 & 0.884$\pm$0.042 & 0.621$\pm$0.049 & 0.813$\pm$0.037 & 0.852$\pm$0.036 \\
MI-Net\citep{deepmil2}           & 0.894$\pm$0.042 & 0.874$\pm$0.043 & 0.630$\pm$0.037 & 0.845$\pm$0.039 & 0.872$\pm$0.032 \\
Attention\citep{deepmil}       & 0.892$\pm$0.040 & 0.858$\pm$0.048 & 0.615$\pm$0.043 & 0.839$\pm$0.022 & 0.868$\pm$0.022 \\
Gated-Attention\citep{deepmil} & 0.900$\pm$0.050 & 0.863$\pm$0.042 & 0.603$\pm$0.029 & 0.845$\pm$0.018 & 0.857$\pm$0.027
\\ \hline 
Prototype Learning & 0.9083$\pm$0.107 & 0.942$\pm$0.070& 0.691$\pm$0.100 & 0.916$\pm$0.0565  & 0.920$\pm$0.060 \\
\hline \\
       \end{tabular}}
    \caption{Result comparison of different approaches.}
    \label{tab:Results}
\end{table}

\subsection{Interpretability}
\paragraph{}As indicated previously, interpretability of the solution is one of the main aspects of prototype learning. To demonstrate this interpretability, here we apply our approach to the MNIST MIL problem which was introduced in \cite{deepmil}. In this case, each instance is an image and each bag consists of images. The task is finding whether a target number exists in images in a bag. To keep things simple, we chose the number of prototypes to be 2. 

\begin{figure}[!h] \label{fig:1}
  \centering
  \begin{subfigure}[b]{0.47\linewidth}
\centering
    \includegraphics[width=\linewidth]{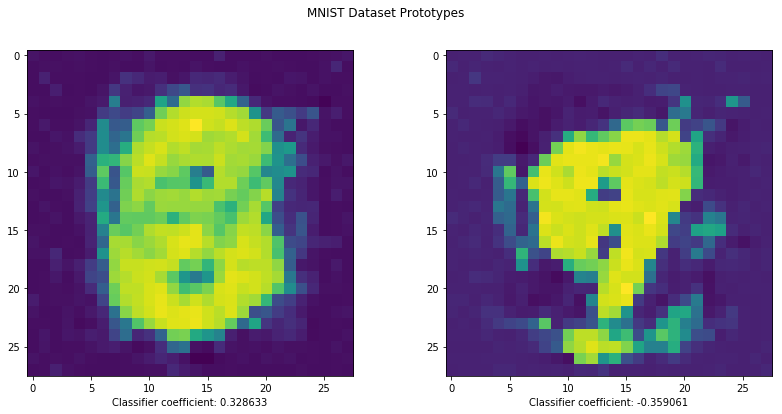}
    \caption{Prototype results when the method is applied to MNIST dataset with finding the 9 problem.}

\label{fig:mnist-1}
  \end{subfigure}
  \rulesep
  \begin{subfigure}[b]{0.47\linewidth}
    \includegraphics[width=\linewidth]{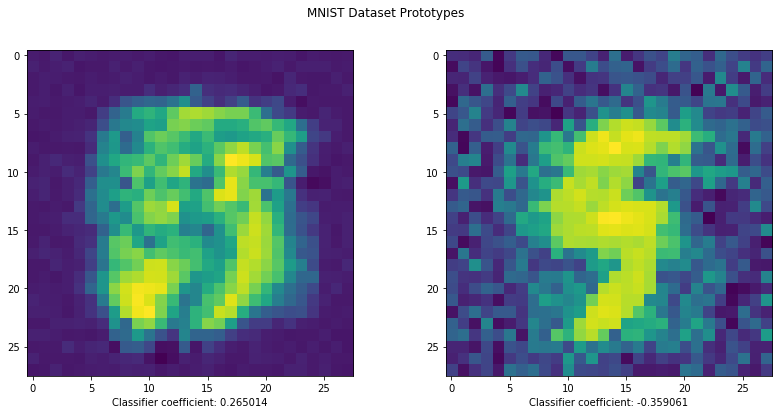}
    \caption{Bags are labeled positive if number 5 is an instance in the bag.}
\label{fig:mnist-2}
  \end{subfigure}
  \caption{MNIST Prototypes.}
  \label{fig:mnist}
\end{figure}

\paragraph{} Examples of prototypes from 2 different runs can be seen in Figure \ref{fig:mnist}. In this application, we only used $min$ as the aggregator for better interpretation. For instance, looking at Figure \ref{fig:mnist-1}, we see that second prototype looks a lot like a 9, and the classifier found a negative coefficient for minimum distance to this prototype. This indicates, if the minimum distance to this prototype is larger the output probability will suffer. Moreover, since the first prototype has a positive coefficient, if the minimum distance of the bag to this prototype is larger, the output probability will be higher. Same analysis can be done for Figure \ref{fig:mnist-2}. 
       
\section{Conclusion}\label{conclusion}
This work presents a prototype learning framework for MIL problems. This simple architecture can be applied to data from all kinds of domains and offers interpretability of the solutions. Although the method is applied to the classical MIL datasets, its modification to different problems is an interesting research direction. Our aim is to present the simplicity of the approach and hence sticked to the logistic regression classifiers and Euclidean distance. However, considering the flexibility of the architecture, one could incorporate more complex classifiers, different distance metrics and different aggregation procedures to obtain more powerful models.

\bibliographystyle{unsrtnat}
\bibliography{references}

\begin{thebibliography}{30}
\providecommand{\natexlab}[1]{#1}
\providecommand{\url}[1]{\texttt{#1}}
\expandafter\ifx\csname urlstyle\endcsname\relax
  \providecommand{\doi}[1]{doi: #1}\else
  \providecommand{\doi}{doi: \begingroup \urlstyle{rm}\Url}\fi

\bibitem[Dietterich et~al.(2001)Dietterich, H.~Lathrop, and
  Lozano-Pérez]{dietterich}
Thomas Dietterich, Richard H.~Lathrop, and Tomás Lozano-Pérez.
\newblock Solving the multiple instance problem with axis-parallel rectangles.
\newblock \emph{Artificial Intelligence}, 89:\penalty0 31--71, 03 2001.

\bibitem[Weidmann et~al.(2003)Weidmann, Frank, and Pfahringer]{weidmann_et_al}
Nils Weidmann, Eibe Frank, and Bernhard Pfahringer.
\newblock A two-level learning method for generalized multi-instance problems.
\newblock \emph{Machine Learning: ECML}, pages 468--479, 2003.

\bibitem[Li et~al.(2013)Li, Tax, Duin, and Loog]{classifier-combining}
Yan Li, David~M.J. Tax, Robert~P.W. Duin, and Marco Loog.
\newblock Multiple-instance learning as a classifier combining problem.
\newblock \emph{Pattern Recognition}, 46\penalty0 (3):\penalty0 865 -- 874,
  2013.

\bibitem[Wang and Zucker(2000)]{lazy_learning}
Jun Wang and Jean-daniel Zucker.
\newblock Solving the multiple-instance problem: A lazy learning approach.
\newblock \emph{Proc. 17th International Con. on Machine Learning}, pages
  1119--1126, 01 2000.

\bibitem[Xu and Frank(2004)]{xu_and_frank}
X.~Xu and E.~Frank.
\newblock Logistic regression and boosting for labeled bags of instances.
\newblock \emph{Proceedings of the Eighth Pacific-Asia Conference on Knowledge
  Discovery and Data Mining}, pages 272--281, 2004.

\bibitem[Xu(2003)]{xu}
X.~Xu.
\newblock Statistical learning in multiple instance problems (m.sc. thesis),
  university of waikato.
\newblock \emph{University of Waikato}, 2003.

\bibitem[Raykar et~al.(2008)Raykar, Krishnapuram, Bi, Dundar, and
  Rao]{raykar_et_al}
V.C. Raykar, B.~Krishnapuram, J.~Bi, M.~Dundar, and R.B. Rao.
\newblock Bayesian multiple instance learning: Automatic feature selection and
  inductive transfer.
\newblock \emph{Proceedings of the 25th International Conference on Machine
  Learning, ACM, New York, NY}, pages 808--815, 2008.

\bibitem[Amores(2013)]{amores}
J.~Amores.
\newblock Multiple instance classification: Review, taxonomy and comparative
  study.
\newblock \emph{Artificial Intelligence 201}, pages 81--105, 2013.

\bibitem[Zhou et~al.(2005)Zhou, Jiang, and Li]{zhou_and_jiang}
Z.~Zhou, K.~Jiang, and M.~Li.
\newblock Multi-instance learning based web mining.
\newblock \emph{Applied Intelligence}, 22(2):\penalty0 135--147, 2005.

\bibitem[Cheplygina et~al.(2013)Cheplygina, Tax, and
  Loog]{cheplygina2013combining}
Veronika Cheplygina, David Tax, and Marco Loog.
\newblock Combining instance information to classify bags.
\newblock \emph{Multiple Classifier Systems, submitted}, 05 2013.

\bibitem[Cheplygina et~al.(2016)Cheplygina, Tax, and Loog]{dissimilarity}
V.~Cheplygina, D.~M.~J. Tax, and M.~Loog.
\newblock Dissimilarity-based ensembles for multiple instance learning.
\newblock \emph{IEEE Transactions on Neural Networks and Learning Systems},
  27\penalty0 (6):\penalty0 1379--1391, June 2016.

\bibitem[Gehler and Chapelle(2007)]{gehler}
Peter Gehler and Olivier Chapelle.
\newblock Deterministic annealing for multiple-instance learning.
\newblock 2007.

\bibitem[Ilse et~al.(2018)Ilse, Tomczak, and Welling]{deepmil}
Maximilian Ilse, Jakub Tomczak, and Max Welling.
\newblock Attention-based deep multiple instance learning.
\newblock 80:\penalty0 2127--2136, 10--15 Jul 2018.
\newblock URL \url{http://proceedings.mlr.press/v80/ilse18a.html}.

\bibitem[Wang et~al.(2018)Wang, Yan, Tang, Bai, and Liu]{deepmil2}
Xinggang Wang, Yongluan Yan, Peng Tang, Xiang Bai, and Wenyu Liu.
\newblock Revisiting multiple instance neural networks.
\newblock \emph{Pattern Recogn.}, 74\penalty0 (C):\penalty0 15--24, February
  2018.
\newblock ISSN 0031-3203.
\newblock \doi{10.1016/j.patcog.2017.08.026}.
\newblock URL \url{https://doi.org/10.1016/j.patcog.2017.08.026}.

\bibitem[{Wu} et~al.(2015){Wu}, {Yinan Yu}, {Chang Huang}, and {Kai
  Yu}]{classification}
J.~{Wu}, {Yinan Yu}, {Chang Huang}, and {Kai Yu}.
\newblock Deep multiple instance learning for image classification and
  auto-annotation.
\newblock In \emph{2015 IEEE Conference on Computer Vision and Pattern
  Recognition (CVPR)}, pages 3460--3469, June 2015.
\newblock \doi{10.1109/CVPR.2015.7298968}.

\bibitem[Tang et~al.(2018)Tang, Wang, Bai, Shen, Bai, Liu, and Yuille]{object1}
Peng Tang, Xinggang Wang, Song Bai, Wei Shen, Xiang Bai, Wenyu Liu, and Alan~L.
  Yuille.
\newblock {PCL:} proposal cluster learning for weakly supervised object
  detection.
\newblock \emph{CoRR}, abs/1807.03342, 2018.
\newblock URL \url{http://arxiv.org/abs/1807.03342}.

\bibitem[Wan et~al.(2019)Wan, Liu, Ke, Ji, Jiao, and Ye]{object2}
Fang Wan, Chang Liu, Wei Ke, Xiangyang Ji, Jianbin Jiao, and Qixiang Ye.
\newblock {C-MIL:} continuation multiple instance learning for weakly
  supervised object detection.
\newblock \emph{CoRR}, abs/1904.05647, 2019.
\newblock URL \url{http://arxiv.org/abs/1904.05647}.

\bibitem[Dennis et~al.(2018)Dennis, Pabbaraju, Simhadri, and Jain]{resource}
Don~Kurian Dennis, Chirag Pabbaraju, Harsha~Vardhan Simhadri, and Prateek Jain.
\newblock Multiple instance learning for efficient sequential data
  classification on resource-constrained devices.
\newblock In \emph{Proceedings of the 32Nd International Conference on Neural
  Information Processing Systems}, NIPS'18, pages 10976--10987, USA, 2018.
  Curran Associates Inc.
\newblock URL \url{http://dl.acm.org/citation.cfm?id=3327546.3327753}.

\bibitem[Angelidis and Lapata(2017)]{sentiment}
Stefanos Angelidis and Mirella Lapata.
\newblock Multiple instance learning networks for fine-grained sentiment
  analysis.
\newblock \emph{CoRR}, abs/1711.09645, 2017.
\newblock URL \url{http://arxiv.org/abs/1711.09645}.

\bibitem[McFee et~al.(2018)McFee, Salamon, and Bello]{sound_event}
Brian McFee, Justin Salamon, and Juan~Pablo Bello.
\newblock Adaptive pooling operators for weakly labeled sound event detection.
\newblock \emph{CoRR}, abs/1804.10070, 2018.
\newblock URL \url{http://arxiv.org/abs/1804.10070}.

\bibitem[Cheplygina et~al.(2015)Cheplygina, Tax, and Loog]{bag_dissimilarity}
Veronika Cheplygina, David~M.J. Tax, and Marco Loog.
\newblock Multiple instance learning with bag dissimilarities.
\newblock \emph{Pattern Recognition}, 48\penalty0 (1):\penalty0 264 -- 275,
  2015.

\bibitem[Ba et~al.(2016)Ba, Kiros, and Hinton]{layernorm}
Lei~Jimmy Ba, Jamie~Ryan Kiros, and Geoffrey~E. Hinton.
\newblock Layer normalization.
\newblock \emph{CoRR}, abs/1607.06450, 2016.
\newblock URL \url{http://arxiv.org/abs/1607.06450}.

\bibitem[Ioffe and Szegedy(2015)]{batchnorm}
Sergey Ioffe and Christian Szegedy.
\newblock Batch normalization: Accelerating deep network training by reducing
  internal covariate shift.
\newblock In \emph{Proceedings of the 32Nd International Conference on
  International Conference on Machine Learning - Volume 37}, ICML'15, pages
  448--456. JMLR.org, 2015.
\newblock URL \url{http://dl.acm.org/citation.cfm?id=3045118.3045167}.

\bibitem[Paszke et~al.(2017)Paszke, Gross, Chintala, Chanan, Yang, DeVito, Lin,
  Desmaison, Antiga, and Lerer]{pytorch}
Adam Paszke, Sam Gross, Soumith Chintala, Gregory Chanan, Edward Yang, Zachary
  DeVito, Zeming Lin, Alban Desmaison, Luca Antiga, and Adam Lerer.
\newblock Automatic differentiation in pytorch.
\newblock 2017.

\bibitem[Kingma and Ba(2015)]{adamopt}
Diederik~P. Kingma and Jimmy Ba.
\newblock Adam: {A} method for stochastic optimization.
\newblock In \emph{3rd International Conference on Learning Representations,
  {ICLR} 2015, San Diego, CA, USA, May 7-9, 2015, Conference Track
  Proceedings}, 2015.
\newblock URL \url{http://arxiv.org/abs/1412.6980}.

\bibitem[Andrews et~al.(2003)Andrews, Hofmann, and
  Tsochantaridis]{andrews2003support}
S.~Andrews, T.~Hofmann, and I.~Tsochantaridis.
\newblock Support vector machines for multiple-instance learning.
\newblock \emph{Advances in Neural Information Processing Systems 15}, pages
  561--568, 2003.

\bibitem[Gartner et~al.(2002)Gartner, Flach, Kowalczyk, and
  Smola]{positive-negative}
T~Gartner, PA~Flach, A~Kowalczyk, and AJ~Smola.
\newblock \emph{Multi-Instance Kernels}, pages 179 -- 186.
\newblock Morgan Kaufmann, 7 2002.

\bibitem[Zhang and Goldman(2002)]{zhang_and_goldman}
Q.~Zhang and S.~Goldman.
\newblock Em-dd: An improved multiple-instance learning technique.
\newblock \emph{Advances in Neural Information Processing Systems}, 14, 2002.

\bibitem[Zhou et~al.(2009)Zhou, Sun, and Li]{zhou}
Z.H. Zhou, Y.Y. Sun, and Y.F. Li.
\newblock Multi-instance learning by treating instances as non-iid samples.
\newblock \emph{Proceedings of International Conference on Machine Learning
  (ICML)}, pages 1249--1256, 01 2009.

\bibitem[Wei et~al.(2017)Wei, Wu, and Zhou]{wei2017scalable}
Xiu-Shen Wei, Jianxin Wu, and Zhi-Hua Zhou.
\newblock Scalable algorithms for multi-instance learning.
\newblock \emph{IEEE transactions on neural networks and learning systems},
  28\penalty0 (4):\penalty0 975--987, 2017.

\end{thebibliography}
\begin{appendices}
\section{Method Formulation} \label{appendix}
Below we give a formulation for the discussed method. We will only demonstrate the formulation for $min$ features for simplicity, which could easily be generalized to any combination of $min$,$max$ and $mean$. Also below we use sigmoid for binary classification, but the same framework could easily be extended to multinomial case utilizing a softmax function.\\
\newline
$D:$ Number of prototypes \\
$K_{i}: $ Number of instances in bag $i$ \\
$L: $ Number of features in any instance \\
$P: $ set of prototypes, $P_{d}$:  $d^{th}$ prototype, a vector with length $L$ \\
$X_i: $ set of instances in bag $i$, $X_{ik}$: $k_{th}$ instance in bag $i$ \\
$y_i: $ Label, class of the bag $i$. \\
$Dist(v_1, v_2): $ Distance between vectors $v_1$ and $v_2$ \\
$\beta_{d}: $ Weight in linear classifier that corresponds to $d_{th}$ prototype \\
$\Phi_{id}$:  $d^{th}$ element of the output vector for bag $i$. Corresponds to the minimum distance of bag $i$ to prototype $d$ after layer normalization. \\
$\sigma(Y) = \frac{1}{1 + e^{-Y}}$ is the sigmoid function \\
$\Lagr_{ce}(y_i, \hat{y}_i): $ Cross-entropy loss where $y_i$ is the label and $\hat{y}_i$ is the prediction \\
$\lambda_{w}: $ Regularization parameter for linear classifier weights \\
$\lambda_{p}: $ Regularization parameter for prototypes \\
$\lambda_{d}: $ Regularization parameter for the extracted distances \\

\begin{equation}\label{eq:objective}
    \!\min_{P, \beta}  \sum_{i=1}^{N}\Lagr_{ce}(y_i, \hat{y}_i) + \lambda_{w}\norm{\beta}_{1} + \lambda_{p}\sum_{d=1}^{D}\norm{P_{d}}_{2} +
    \lambda_{d}\sum_{i=1}^{N}\sum_{d=1}^{D}\Phi_{id}
\end{equation}

\begin{equation} \label{eq:distances}
\Phi_{id} = \min_{X_{ik} \in X_i}Dist(X_{ik}, P_d)
\end{equation}

\begin{equation} \label{eg:prediction}
    \hat{y}_i = \sigma (\beta_0 + \sum_{d=1}^{D}\beta_d \Phi_{id})
\end{equation}

The objective function given in Equation \ref{eq:objective} is optimized over the prototypes, $P$, and the linear classifier weights, $\beta$. Different lea rning rates are used for prototypes and the classifier parameters, namely $\alpha_{w}$ for weights and $\alpha_p$ for prototypes. The model is trained for 100 epochs, minibatch size 1 is used with Adam optimizer \citep{adamopt} .\\

We also perform layer normalization. Namely, each row in the transformed distance space is scaled to zero mean and unit variance as illustrated in Equations \ref{eq:4a}, \ref{eq:4b}. Layer normalization is important because it stabilizes the issues that could occur during optimization due to the scale of distance features and it reduces the sensibility of linear classifier to the scale of distances but it keeps the relative distance information. As a side benefit, we observe that as argued in \citep{layernorm}, it speeds up the convergence.

\begin{subequations}
    \begin{equation}\label{eq:4a}
    \mu_{i} = \frac{1}{D}\sum_{d=1}^D\Phi_{id}, \sigma_i = \sqrt{\frac{1}{D}(\Phi_{i}-\mu_i)^2}
    \end{equation}
    \begin{equation}\label{eq:4b}
    \Phi_{i} = \frac{\Phi_{i} - \mu_i}{\sigma_i}
    \end{equation}
    \label{eq:4}
\end{subequations}

The parameters of our experiments are as the following: $D=24$, $\lambda_p=0.05$, $\lambda_w=0.05$, $Dist$ calculates Euclidean distance, $\alpha_p=0.0001$, $\alpha_w=0.00005$. In practice, any differentiable distance metric could be used in this framework with gradient based optimization methods.

\end{appendices}
\end{document}